\documentclass[conference]{IEEEtran}

\usepackage[margin=0.75in]{geometry} 
\usepackage{amsmath,amsthm,amssymb}
 \usepackage{graphicx}

\theoremstyle{remark}

\usepackage{stackengine}

\ifCLASSINFOpdf
\else
\fi
\begin{document}
%
\title{Lung Cancer Detection and Classification based on Image Processing and Statistical Learning}

\author{\IEEEauthorblockN{Md Rashidul Hasan}
\IEEEauthorblockA{Department of Mathematics and Statistics\\
The University of New Mexico\\Albuquerque, New Mexico, USA}
\and
\IEEEauthorblockN{Muntasir Al Kabir}
\IEEEauthorblockA{Anderson Department of Marketing, Information Systems, \\Information Assurance, and Operations Management\\
The University of New Mexico\\Albuquerque, New Mexico, USA}}


%


\maketitle

\begin{abstract}
Lung cancer is one of the death threatening diseases among
human beings. Early and accurate detection of lung cancer
can increase the survival rate from lung cancer. Computed
Tomography (CT) images are commonly used for detecting
the lung cancer.Using a data set of thousands of high-resolution lung scans collected from Kaggle competition [1], we will develop algorithms that accurately determine  in the lungs are cancerous or not. The proposed system promises better result than the existing systems, which would be beneficial for the radiologist for the accurate and early detection of cancer. The method has been tested on 198
slices of CT images of various stages of cancer obtained
from Kaggle dataset[1] and is found satisfactory results. The accuracy of the proposed method in this dataset is 72.2\%
\end{abstract}


%
\IEEEpeerreviewmaketitle

\section{Introduction}
Lung cancer is one kind of decease that grows uncontrolled way and 
form abnormal cells in the lung. These cells do not function like other 
normal cells. Because of DNA mutation by different factors like smoking, air 
pollution, Inherited gene changes, cancer can grow in human lungs. According to American Cancer Society[1], among all new cancers about 14\% are lung cancers.They also estimate in 2018, there are about 234,030 new lung cancer in United States and about 154,050 deaths because of lung cancer. Now a days, the reason of death is far beyond than prostate, colon, and breast cancers combined to lung cancer.  
\\

Objective of this study is to detect lung cancer using image processing techniques. CT scanned lung
images of cancer patients are acquired from Kaggle Competition dataset. Using image processing techniques like preprocessing, Segmentation and feature extraction, area of interest is separated. Developing the algorithm, features like area, perimeter and entropy are extracted from all the images. The parameter values obtained from these features
are compared with the normal values suggested by a physician. From the comparison result, cancer noodles is detected. A graphical user interface is developed to scan all the images and display the features and cancer
noddles. This system can help in early detection of lung cancer more accurately.
\\

Shojaii et.el (2005) [5] presented lung segmentation technique using watershed transform along with internal and external marker. They also used rolling ball filter for the smoothing of the contour and to fill the cavities of the cancer noodles.
\\ 
Nivetha et.el (2014) [6] used genetic algorithm to select particular features and GLCM for the extraction. They used Support vector machines (SVM) to classify stages of lung cancer.
\\
Kaur et.el (2013) [7] investigated several existing literature and identify the bilateral filter is best solution in using marker based watershed segmentation algorithm.
\\
Al-Fahoum et.el (2014) [8] proposed a computer aided detection(CAD) system to detect the lung cancer areas from CT images.
\\ 
Ignatius et.el (2015) [9] Used classifier such as SVM, NBM, NB Tree and Random Tree on 200 slices of CT images of several stages of cancer and found 94.4\% accuracy while using Random Tree.
\\
Bush (2017) [10] presented ResNet CNN model to identify and localized different noodles. However this model does not for localizing exact position noodle.
\\
Ciumpi et.el (2017) [11] applied  a deep learning system to different dataset, one from Italian MILD screening trail as training data and another from the Danish DLCST screening trial as test data of lung cancer patients to compare the difference between computer and human as a observer.
\\
Nayayanan et.el (2018) [12] used three different training data based on sliced thickness in computer aided detection.

\section{MATERIALS \& METHODS}

\begin{figure}[!htbp]
\centering
\includegraphics[width=2.5in]{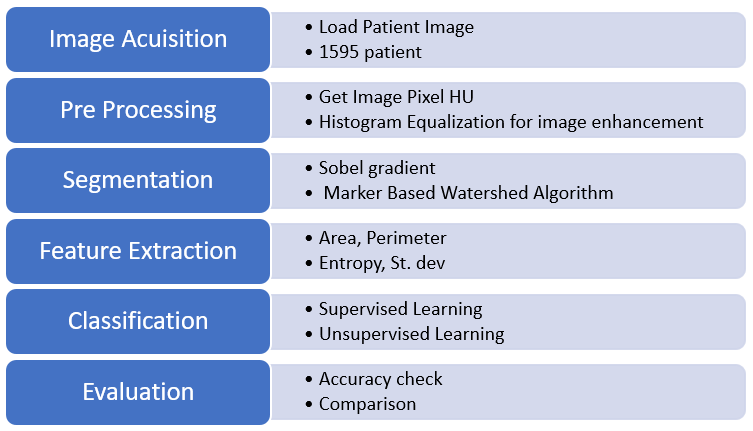}
\caption{CT Image analysis steps for breast cancer detection}
\label{pr}
\end{figure}

In figure \ref{pr} step by step procedures for CT image analysis is shown which will be discussed in details in the following
sections.

\subsection{Image Processing}

\subsubsection{Image Aquisition}
Because of low noise and better clarity, CT scan images of Lung cancer patient are more useful compared to MRI and X-ray.  Due to its lesser distortion property, CT scan is easier to handle for the preprocessing part. For our research work, the CT images has been acquired from Kaggle competition dataset. Now a days, DICOM (Digital Imaging and Communication in Medicine) is a standard format for medical imaging. Figure
\ref{or} shows a typical CT image of lung cancer patient used for analysis. The acquired images are in the raw form and observed a lot of noise. 

\subsubsection{Pre-Processing}
\paragraph{Smoothing}
To improve the contrast, clarity, separate the background noise, it is required to pre-process the images. Hence, various techniques like smoothing, enhancement are applied to get image in required form. It suppresses the noise or other small fluctuations in the image; equivalent to the suppressions of high frequencies in the frequency domain. Smoothing also blurs all sharp edges that bear important information about the image. To remove the noise from the images, median filtering is used. Median filtering is a non-linear operation often used in image processing to reduce salt and pepper noise. In general, the median filter allows a great deal of high spatial frequency detail to pass while remaining very effective at removing noise on images where less than half of the pixels in a smoothing neighborhood have been affected. B=medfilt2(A,[m,n]) performs median filtering of the matrix A in two dimensions. Each output pixel contains the median value in the
m x n neighborhood around the corresponding pixel in the image. 

\begin{figure}[!htb]\centering
   \begin{minipage}{0.24\textwidth}
     \frame{\includegraphics[width=\linewidth]{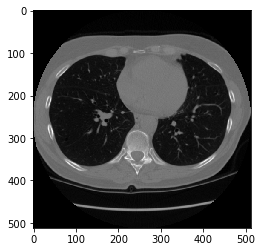}}
     \caption{Orginal Image}
     \label{or}
   \end{minipage}
   \begin {minipage}{0.24\textwidth}
     \frame{\includegraphics[width=\linewidth]{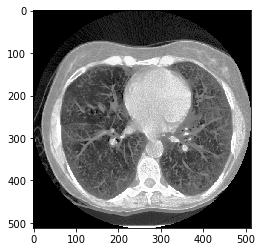}}
     \caption{Enhanced Image}
     \label{ep}
   \end{minipage}
\end{figure}

\paragraph{Enhancement}
Enhancement technique is used to improve the interpretability or perception of information in images for human viewers, or to provide better input for other automated image processing techniques. Image enhancement can be classified in two main categories, spatial domain and frequency domain. Here histogram equalization is used for enhancement purpose and the output after performing enhancement from original image is shown in figure \ref{ep}.

\subsubsection{segmentation}
Image segmentation is a process of subdividing an image into the constituent parts or objects in the image. So the main purpose of subdividing an image into its constituent parts or objects present in the image is that we can further analyze each of the constituents or each of the objects present in the image once they are identified or we have subdivided them. So resulted output of image segmentation is a collections of segment of entire image. In our method we use marker-controlled watershed segmentation.

\begin{figure}[!htb]\centering
   \begin{minipage}{0.24\textwidth}
     \frame{\includegraphics[width=\linewidth]{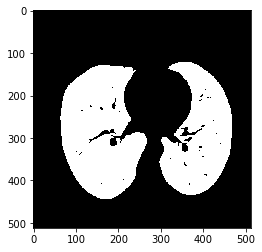}}
     \caption{Internal Marker}
     \label{im}
   \end{minipage}
   \begin {minipage}{0.24\textwidth}
     \frame{\includegraphics[width=\linewidth]{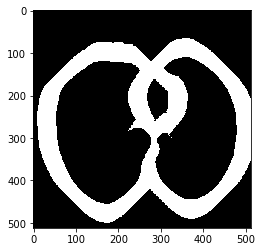}}
     \caption{External Marker}\label{em}
   \end{minipage}
   \begin {minipage}{0.24\textwidth}
     \frame{\includegraphics[width=\linewidth]{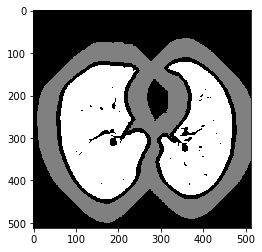}}
     \caption{Watershed Marker}
     \label{wm}
   \end{minipage}   
\end{figure}

\begin{figure}[!htb]\centering
   \begin{minipage}{0.24\textwidth}
     \frame{\includegraphics[width=\linewidth]{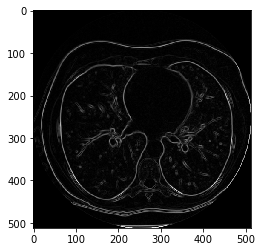}}
     \caption{Sobel Gradient Image}
     \label{sg}
   \end{minipage}
   \begin {minipage}{0.24\textwidth}
     \frame{\includegraphics[width=\linewidth]{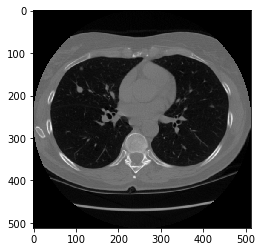}}
     \caption{Segmented Lung Image}
     \label{sl}
   \end{minipage}
\end{figure}

In order to use marker based watershed segmentation, we use internal marker shown in figure \ref{im}, that is definitely lung tissue and an external marker shown in figure \ref{em}. to find the precise border of the lung we also used the Sobel-Gradient-Image shown in figure \ref{sg} of our original scan.

The different steps involved in Marker Controlled Segmentation [2] are the following:
\\Step 1: Read in the color image and convert it to gray scale image.
\\Step 2: Compute the Gradient Magnitude as the segmentation function.
\\Step 3: Mark the foreground objects within the image.
\\Step 4: Find out the background marker points within the image.
\\Step 5: Find out the watershed transform of the segmented function of the image.
\\Step 6: Resultant segmented binary image shown in figure \ref{sl}  is obtained.

\subsubsection{Feature Extraction}
This stage is an important stage that uses algorithms and techniques to detect and isolate various desired portions or shapes segmented image. After labeling the segmented image we extracted the various features.  The basic characters of feature are area, perimeter and eccentricity. These are measured in scalar.
These features are defined as follows:
\\\textbf{Area} is one of the key parameters required for classification
process. Area actually tells us about the size of the lump.
\\\textbf{Perimeter}, another important parameter gives us the idea about
the boundary of the defected cell.
\\\textbf{Standard Deviation}, σ is the estimate of the mean square deviation of the grey scale pixel value from its mean, µ.
\\\textbf{Skewness} characterizes the degree of asymmetry of a pixel distribution in the specified ROI around its mean.
\\\textbf{Kurtosis} measures the peakness or flatness of a distribution relative to a normal distribution.
\\\textbf{Entropy} is a measure of the maximal amount of potential information given by the segmented ROI.

\section{Classification}
\begin{figure}[!htpb]
\centering
\includegraphics[width=8cm,height=4cm]{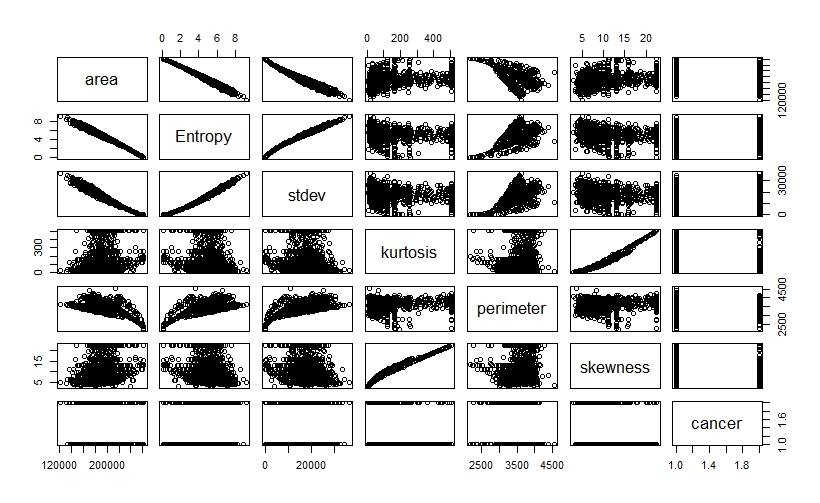}
\caption{Scatter plot matrix of training data set}
\label{fig:scatter}
\end{figure}

Before discussing the classification, we divide our data set into training and test data. The training data set consists of 1397 patients where 1035 patients do not have cancer and rest of 362 do have. On the other hand, our test data set contains 198 patients where 57 patients are carrying cancerous region and 141 without that region. In this section, We want to choose a model based on our training data and then test the model for accuracy. For choosing the model we tried both supervised and unsupervised learning. 
\\\\However, Our goal is to predict the response variable cancer (yes or no) which is a categorical variable. Moreover, We  want to make sure that there is no problem of collinearity among the predictor variables. Figure \ref{fig:scatter} shows the scatter plot matrix of all variables to check the collinearity. This matrix plot indicates high correlation in between the predictors area, entropy, and standard deviation. Furthermore, skewness and kurtosis are also highly correlated to each other. Therefore this collinearity suggests us to eliminate some of our predictor variables.
We used best subset selection method for eliminating non significant predictors. Then we applied different supervised and unsupervised learnings. The methods and classifications are discussed below: 
\subsection{Best Subset selection}
\begin{figure}[!htpb]
\centering
\includegraphics[width=8cm,height=4cm]{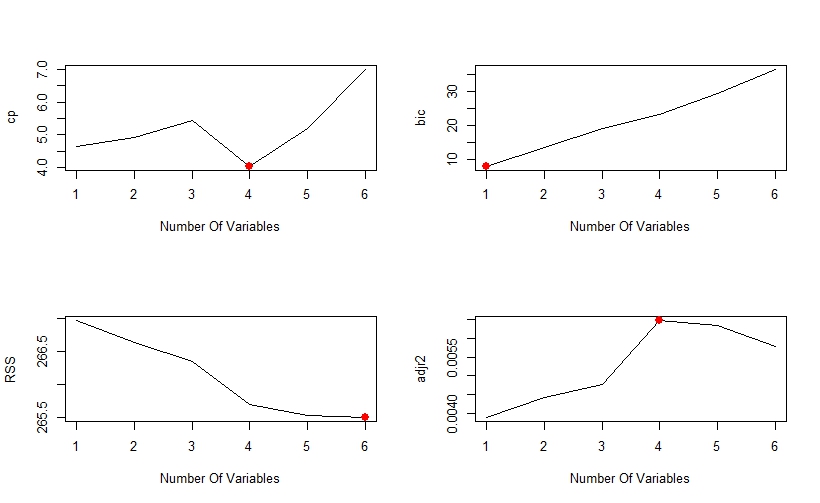}
\caption{Plot for $C_p$, BIC, RSS, and $adjR^2$}
\label{fig:cpbicrssadr2}
\end{figure}

We ran a linear regression model for each possible combination of the X's. Fitting all models with $k$ predictors where $k = 0, 1,....,p$, we selected the best model, $M_k$, using minimum RSS, $C_p$, BIC, or
highest adjusted $R^2$.
\\\\For our data set both $C_p$ and adjusted $R^2$ suggested 4 predictors as shown in figure \ref{fig:cpbicrssadr2}. 
Then we tried four as well as three predictors separately and found that entropy, standard deviation and perimeter are statistically significant. However, for classification we tried two cases (i) all predictors and (ii) three predictors to see if there were any improvisation in accuracy level. We applied multiple logistic regression in the next section. 
\subsection{Multiple Logistic Regression}
For various predictors $X_1, X_2,.....,X_p$, the multiple logistic regression is
generalized as follows:
$$P(x)=\frac{e^{\beta_0+\beta_1X_1+......+\beta_pX_p}}{1+e^{\beta_0+\beta_1X_1+......+\beta_pX_p}}$$
where $X=(X_1,.....X_p)$ are $P$ predictors. 
\\\\Using all the predictors, this logistic regression method gave us no significant predictor variables except the standard deviation. However, this method predicted 60.1\% data accurately. With the three predictors logistic regression model then gave us a improved accuracy level of 69.19\%. Next, section applied linear discriminant analysis.
\subsection{Linear Discriminant Analysis (LDA) for $p>1$}
\begin{figure}[!htpb]
\centering
\includegraphics[width=8cm,height=4cm]{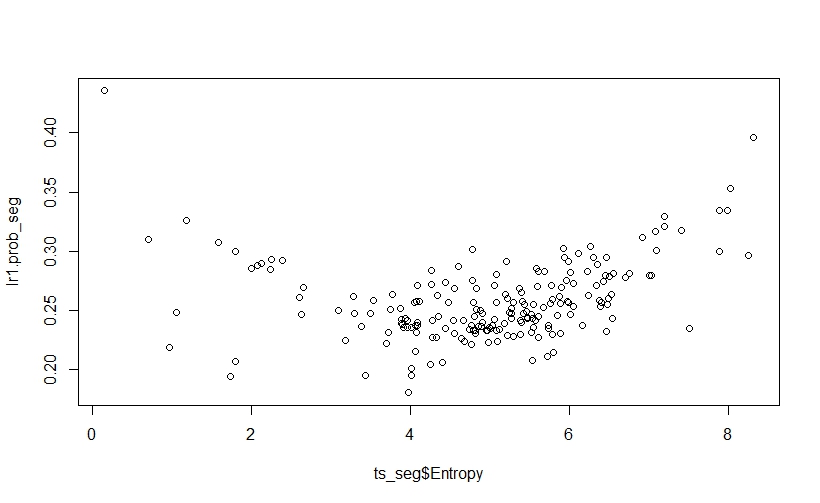}
\caption{relation between cancer and entropy}
\label{fig:lda}
\end{figure}

Assume that $X = (X_1, X_2, . . . , X_p)$ is drawn from a multivariate normal distribution, with a class-specific multivariate mean vector and a common covariance matrix. Therefore,
$$f(X) = \frac{1}{
(2\pi)^{p/2}\mid\sum\mid^{1/2}}exp (-\frac{1}{2}(x - \mu)^T (\sum)^{−1}(x-\mu)) $$
Then the Bayes classifier assigns an observation $X = x$ to the class for which
$$\delta_k(x)=x^T(\sum)^{-1}\mu_k-\frac{1}{2}\mu_k^T(\sum)^{-1}\mu_k+log(\pi_k)$$
Unfortunately, this method did not work. One of the reasons might be the relationship between the response and predictors are not linear. For example, figure \ref{fig:lda} shows the curvilinear relation between cancer and  entropy. Next, we applied quadratic discriminant analysis. 

\subsection{Quadratic Discriminant Analysis (QDA)}
Again assume that $X = (X_1, X_2, . . . , X_p)$ is drawn from a multivariate normal distribution. Therefore, the Bayes classifier assigns an observation $X = x$ to the class for which
\\$\delta_k(x)=-\frac{1}{2}x^T(\sum)^{-1}x+x^T(\sum)^{-1}\mu_k-\frac{1}{2}\mu_k^T(\sum)^{-1}\mu_k-\frac{1}{2}log\mid(\sum)_k\mid+log(\pi_k)$
\\\\For QDA, all predictor variables gave us 69.69\% accuracy and when used three predictors we got slightly higher accuracy level of 71.21\%. Next, we applied  K-nearest neighbors Regression.
\subsection{K-nearest neighbors (KNN) Regression}
To predict Y for a given X value, consider the K closest
points to X in training data and take the average of the
response,
$$f(X) = \frac{1}{k}\sum_{X_i\in N_i}Y_I$$
$N_i$ is the set of $K$ neighbors based on $X$
\\\\For KNN, All predictor variables gave us 62.12\% accuracy and when used three predictors we got slightly higher accuracy level of 64.64\%. Next, we applied classification trees.
\subsection{Classification Trees}
\begin{figure}[!htpb]
\centering
\includegraphics[width=8cm,height=4cm]{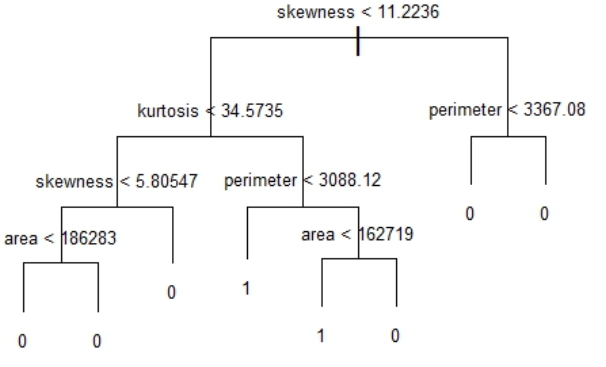}
\caption{Classification tree using 8 nodes}
\label{fig:tree}
\end{figure}

A classification tree is used to predict a qualitative response rather than a quantitative one. For a classification tree, we predict that each observation belongs to the most commonly occurring class of training observations in the region to which it belongs. In interpreting the results of a classification tree, we are often interested not only in the class prediction corresponding to a particular terminal node region, but also in the class proportions among the training observations that fall into that region.
\\\\A large tree with lots of leaves tends to overfit the training data. We may consider to reduce the tree by ”pruning” some of the leaves. This can lead to improved accuracy.
Use cross-validation to check which tree has the lowest RSS or error rate. Fortunately, there is software in place to perform all these calculations. 
\\\\Because of some computational complexity we could not use all the training data for classification trees. However, we managed to handle 600 observations. Then  all predictor variables gave us 71.71\% accuracy with 8 nodes as shown in figure \ref{fig:tree} and after pruning, 3 nodes had been used but we got exactly same accuracy level.  
Next, we applied random forest method.
\subsection{Random Forests}
Random forests is a very efficient statistical learning method. It builds on bagging (in bagging, we build a number forest of decision trees on bootstrapped training samples.), but provides an improvement because it de-correlates the trees.Build a number of decision trees on bootstrapped training samples. Each time a split in a tree is considered, a random sample of $m$ predictors is chosen as split candidates from the full set of p predictors, typically $m = \sqrt{p}$.

\begin{figure}[!htpb]
\centering
\includegraphics[width=8cm,height=4cm]{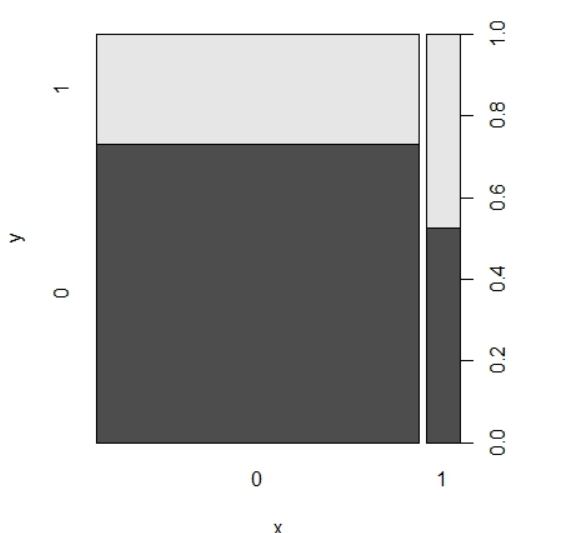}
\caption{Classification using random forest}
\label{fig:forest}
\end{figure}

Suppose that there is a very strong predictor. The rest are moderate to poor predictors. For the bagged trees, most of the them will have the strong predictor for the first split. All bagged trees will look similar and the respective predictions, highly correlated. Averaging highly correlated quantities does not help with variance reduction. Random forests de-correlate the bagged trees.
\\\\Figure \ref{fig:forest} represents the classification when used the random forest. Surprisingly the accuracy level is as same as the classification trees i.e. 71.71\%.
In the next section, we applied support vector machine.
\subsection{Support Vector Machine (SVM)}
\begin{figure}[!htpb]
\centering
\includegraphics[width=8cm,height=4cm]{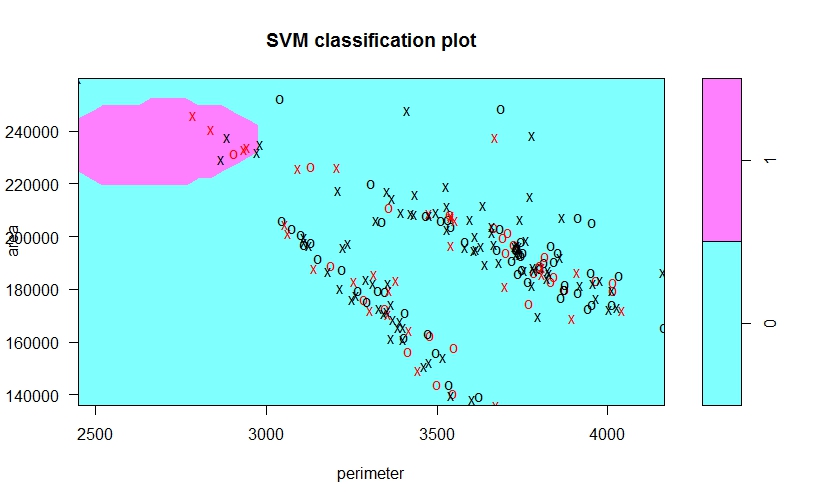}
\caption{Classification using support vector machine}
\label{fig:svm}
\end{figure}

The support vector classifier finds the optimal hyperplane in the space spanned by $X_1, X_2,......,X_p$. In support vector machine we create a transformation $b_1(X), b_2(X),......,b_m(X)$. Then we find the support vector classifier in the transformed space. This produces a linear support classifier in $b_1(X), b_2(X),......,b_m(X)$ but its non linear in the original $X_1, X_2,......,X_p$. 
\\\\In general, a support vector machine can be expressed through kernels as 
$$f(x) = \beta_0+\sum_{i\in S}\alpha_iK(X,X_i)$$
where $K(.,.)$ is a Kernel function between two vectors. We define two vectors as $x_i = (x_{i,1}, x_{i,2},......x_{i,p})$ and $x_l = (x_{l,1}, x_{l,2},.....,x_{l,p})$ then the Possible kernels are (i) inner product kernel is $K(X_i,X_l)=\sum_{j=1}^pX_{i,j}X_{l,j}= <X_i,X_l>$ (ii) polynomial kernel is $K(X_i,X_l)=\sum_{j=1}^p(1+X_{i,j}X_{l,j})^d$, and (iii) radial kernel ($\gamma >0$) is $K(X_i,X_l)=exp(-\gamma\sum_{j=1}^p(1+X_{i,j}X_{l,j})^2)$  
\\\\SVM also gave us 71.71\% before tuning the cost and gamma parameters. Then we tuned these two parameters and got the best results for cost=1 and gamma=1. This moderately improved our accuracy level to 72.22\%. Two predictors, area and perimeter have been used for SVM as shown in figure \ref{fig:svm}.
\\Finally, K-means clustering also applied in the next section.
\subsection{K-Means Clustering}
\begin{figure}[!htpb]
\centering
\includegraphics[width=10cm,height=4cm]{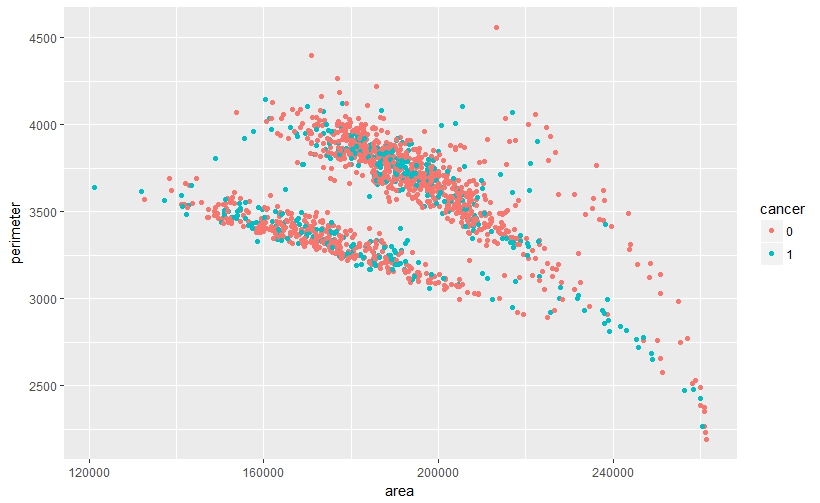}
\caption{Classification using k-means clustering}
\label{fig:kmean}
\end{figure}

K-means clustering is a simple and elegant approach for partitioning a
data set into K distinct, non-overlapping clusters. To perform K-means
clustering, we must fixed the desired number of clusters K. Suppose 
$C_1, C_2,.....,C_K$ are indices of the observations that define
each cluster where $C_k\cap C_{k^\prime} = \varnothing$, i.e. sets are mutually exclusive. Let $W(C_k)$ measures how much observations differ within a
cluster. The goal is to select $C_1, C_2,.....,C_K$ so that they minimize
$$\sum_{i=1}^kW(c_i)$$
where $W(c_k)=\frac{1}{|c_k|}\sum_{i,i^\prime\in c_k}||X_i-X_{i^\prime}||^2$, here $x_i$ is the vector of all covariates for observation $i$, $|C_k|$ is the total number of elements in $C_k$. In this formulation, $W(C_k)$ depends on the mean of each
variable $X_j$ for $C_k$ (centroids).
\\\\When used all predictors k-means clustering for training data gave 52.97\% accuracy and for three predictors we got 54.67\%.  For test data using all predictors gave the accuracy level of 47.47\% and three predictors gave slightly improved level of 55.05\%. Figure \ref{fig:kmean} shows the k-means clustering for area and perimeter. 
\section{Results}
\begin{figure}[!htpb]
\centering
\includegraphics[width=8cm,height=6cm]{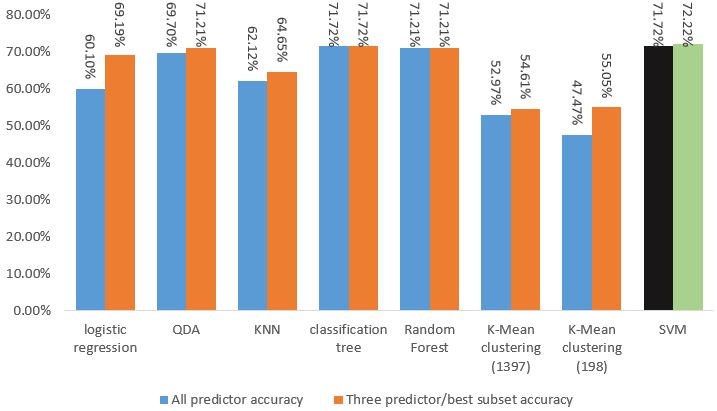}
\caption{Bar graphs comparing the accuracy level}
\label{fig:capture}
\end{figure}

Figure \ref{fig:capture} represents the summary of accuracy level. Blue and orange color indicates the the percentage of accuracy for all predictors and three predictors respectively. In contrast, different colors for SVM is for two different cost and gamma parameters. This bar graph also shows that SVM provides us the highest accuracy level while QDA, classification tree, and random forest are competing with SVM.   


\section{Future Enhancement}
In the proposed system we used only watershed marker based segmentation in image processing part. Future work we want to use some other segmentation technique and compare. We believe that will increase our extracted feature quality. We also considering to use some other filter and image enhancement method. The accuracy can be increased by
extracting more features of the tumor, increasing the size of
the dataset.

\section{Conclusion}
The proposed lung cancer detection identifies the tumor within the lung. The CT image is pre-processed and the pre-processed image is then subjected to segmentation by using Marker Controlled watershed segmentation. Segmented image is used for feature extraction. With the extracted features the tumor is detected within the lung. Both supervised and unsupervised classifier is used for the identifying of the cancer. The accuracy rate of the proposed system is 72.2\% by using support vector machine. Thus this system helps the radiologist to identify the stage of the tumor and increase the accuracy.




\begin{thebibliography}{1}

\bibitem{IEEEhowto:kopka}
https://www.kaggle.com/c/data-science-bowl-2017
\bibitem{IEEEhowto:Iyla}
Ilya Levner, Hong Zhangm ,“Classification driven Watershed segmentation ”, IEEE TRANSACTIONS ON IMAGE PROCESSING VOL. 16, NO. 5, MAY 2007
\bibitem{IEEEhowto:math}
http://in.mathworks.com/help/images/examples/markercontrolled-watershed-segmentation.html
\bibitem{IEEEhowto:bcf}
http://www- bcf.usc.edu/ gareth/ISL/index.html
\bibitem{IEEEhowto:bcf}
https://ieeexplore.ieee.org/document/1530294/
\bibitem{IEEEhowto:bcf}
http://www.rroij.com/open-access/lung-cancer-detection-at-early-stage-usingpetct-imaging-technique.php?aid=45487
\bibitem{IEEEhowto:bcf}
http://www.ijeit.com/Vol
\bibitem{IEEEhowto:bcf}
http://www.sciedu.ca/journal/index.php/jbgc/article/view/3554
\bibitem{IEEEhowto:bcf}
https://pdfs.semanticscholar.org/e8bf/3d6b4d897fd3c9e13feed03636d3ee0f1845.pdf
\bibitem{IEEEhowto:bcf}
https://pdfs.semanticscholar.org/0aad/16fbbe39a9d5703e2ed11b97b08f7285b513.pdf
\bibitem{IEEEhowto:bcf}
https://www.nature.com/articles/srep46479
\bibitem{IEEEhowto:bcf}
https://www.ncbi.nlm.nih.gov/pubmed/29487880






\end{thebibliography}
%

\end{document}